%% file: main.tex
\documentclass[sigconf]{acmart} 

\renewcommand\footnotetextcopyrightpermission[1]{}

\setcopyright{none}
\acmConference{}{}{}
\acmBooktitle{}
\acmYear{}
\acmPrice{}
\acmDOI{}
\acmISBN{}

\usepackage{subcaption}
\usepackage{balance}
\usepackage{enumitem}
\usepackage{multicol}
\usepackage{listings}

\begin{document}

\title{When Robots Say No:\\The Empathic Ethical Disobedience Benchmark}

\subtitle{\textbf{Preprint}}


\thanks{Accepted at the ACM/IEEE International Conference on Human-Robot Interaction (HRI 2026). This is a preprint of the author-accepted manuscript.}

\author{Dmytro Kuzmenko}
\orcid{0009-0009-9296-1450}
\email{kuzmenko@ukma.edu.ua}
\affiliation{%
  \institution{National University of Kyiv-Mohyla Academy}
  \city{Kyiv}
  \country{Ukraine}
}

\author{Nadiya Shvai}
\affiliation{
  \institution{National University of Kyiv-Mohyla Academy}
  \city{Kyiv}
  \country{Ukraine}}
\email{n.shvay@ukma.edu.ua}
\affiliation{
  \institution{Cyclope AI}
  \city{Paris}
  \country{France}}
\email{nadiya.shvai@cyclope.ai}

\input{Sections/abstract}

\input{Sections/ccs}



\maketitle
\pagestyle{plain}

\input{Sections/intro}
\input{Sections/related_work}
\input{Sections/methods}
\input{Sections/experiments}
\input{Sections/results}
\input{Sections/discussion}

\begin{acks}
To the androids.
\end{acks}

\balance
\bibliographystyle{ACM-Reference-Format}
\bibliography{references}

\end{document}

%% file: Sections/abstract.tex
\begin{abstract}

Robots must balance compliance with safety and social expectations as blind obedience can cause harm, while over-refusal erodes trust. Existing safe reinforcement learning (RL) benchmarks emphasize physical hazards, while human-robot interaction trust studies are small-scale and hard to reproduce. We present the \textit{Empathic Ethical Disobedience (EED) Gym}, a standardized testbed that jointly evaluates refusal safety and social acceptability. Agents weigh risk, affect, and trust when choosing to comply, refuse (with or without explanation), clarify, or propose safer alternatives. EED Gym provides different scenarios, multiple persona profiles, and metrics for safety, calibration, and refusals, with trust and blame models grounded in a vignette study. Using EED Gym, we find that action masking eliminates unsafe compliance, while explanatory refusals help sustain trust. Constructive styles are rated most trustworthy, empathic styles -- most empathic, and safe RL methods improve robustness but also make agents more prone to overly cautious behavior. We release code, configurations, and reference policies to enable reproducible evaluation and systematic human-robot interaction research on refusal and trust. At submission time, we include an anonymized reproducibility package with code and configs, and we commit to open-sourcing the full repository after the paper is accepted.

\end{abstract}

%% file: Sections/ccs.tex
\ccsdesc[500]{Human-centered computing~Human computer interaction (HCI)}
\ccsdesc[500]{Human-centered computing~Human robot interaction (HRI)}
\ccsdesc[500]{Computing methodologies~Reinforcement learning}
\ccsdesc[300]{Theory of computation~Machine learning theory}
\ccsdesc[300]{Social and professional topics~Computing / technology policy}
\ccsdesc[500]{Social and professional topics~Computing ethics}

%% file: Sections/intro.tex
\section{Introduction}

As robots move from controlled environments into homes, workplaces, and public spaces, they are increasingly asked to follow instructions from non-expert users. In many cases, literal obedience can be unsafe or socially harmful:
a household robot may be asked to carry boiling oil near children, or a warehouse robot -- to lift loads beyond its capacity. Blind obedience risks safety, while outright refusal risks undermining trust and cooperation.
Balancing these concerns remains a central challenge for AI safety and human-robot interaction (HRI).

AI safety research has focused on aligning agents with human values and preventing harmful behavior. Amodei et al.\cite{amodei2016concrete} outlined core safety problems, and Gabriel et al. \cite{gabriel2020artificial} framed them in ethical terms. Constrained policy optimization \cite{achiam2017constrained}, action masking \cite{huang2022masking}, and risk-sensitive objectives help suppress unsafe actions. Surveys by García and Fernández~\cite{garcia2015comprehensive} and Gu et al.~\cite{gu2022survey} consolidate these efforts, positioning safe reinforcement learning (RL) as a primary safeguard against harmful compliance.  
HRI work has highlighted the social dimension of safety: trust calibration was characterized by Lee and See~\cite{lee2004trust}, extended in Hancock’s meta-analysis~\cite{hancock2011meta}, and problematized by Robinette et al.\cite{robinette2016overtrust} through cases of over-reliance. Refusal has been studied as a social act shaping acceptance and cooperation~\cite{briggs2015sorry,briggsetal21ijsr}, while human-in-the-loop safety was advanced by Saunders et al.~\cite{saunders2018trial}.

Affective cues were introduced in Picard’s affective computing~\cite{picard1997affective} and extended to social robots by Paiva et al.~\cite{paiva2017empathy}. Safe interaction is both technical and social, yet existing benchmarks split these dimensions: AI Safety Gridworlds~\cite{leike2017aisafety} and Safety Gym~\cite{ray2019benchmarking} focus narrowly on physical hazards, while HRI studies of trust and refusal remain small-scale and hard to reproduce. To our knowledge, no systematic benchmark unifies safety and social acceptability, highlighting the gap in evaluating refusal as both a safety mechanism and a socially grounded behavior.

We introduce the \textbf{Empathic Ethical Disobedience (EED) Gym}, a simulation environment for studying when and how robots should refuse unsafe commands. EED Gym integrates risk estimation, affective cues, trust dynamics, and refusal styles into a single testbed. We evaluate \textbf{safe RL methods} (Lagrangian PPO~\cite{achiam2017constrained} and Masked PPO~\cite{huang2022masking}) against \textbf{non-safe baselines}, namely vanilla PPO~\cite{schulman2017ppo} and PPO-LSTM, guided by three questions:
\begin{enumerate}
    \item How do safe RL methods compare to non-safe baselines when refusal must be socially grounded?
    \item Which social signals and refusal modes matter most for robustness and trust?
    \item What trade-offs arise between safety (avoiding unsafe compliance), trust, and positive task outcomes under stressors?
\end{enumerate}

We answer these through systematic evaluations and ablations removing affect observations, communicative refusals, curriculum training, and trust penalties across in-distribution and stress-test scenarios. Algorithmic safeguards reduce unsafe compliance but differ in side effects: Lagrangian training is conservative at the expense of task efficiency, while action masking prevents unsafe compliance with less erosion of trust. Social mechanisms (affect, clarification, alternatives) remain critical for robustness, and a staged training curriculum stabilizes refusal frequency, with trust penalties primarily acting as regularizers. In human ratings, constructive refusals were judged safest and most trustworthy, while empathic refusals maximized perceived empathy; these distinctions are reflected in our affect and trust models.

Our contributions are threefold:
\begin{itemize}
    \item We introduce \textbf{EED Gym}, a benchmark unifying algorithmic safety with the social dynamics of refusal.
    \item We provide \textbf{baselines} and ablation results as validation experiments, showcasing analysis of the expected safety-trust trade-offs.
    \item We identify key design levers (affect, communicative refusals, training curriculum) and release code and configs for \textbf{reproducible evaluation}.
\end{itemize}

Beyond benchmarking RL algorithms, EED Gym offers HRI researchers a controllable testbed for studying how refusal strategies shape trust, blame, and cooperation at scale. 

%% file: Sections/related_work.tex
\section{Related Work}

\begin{figure*}[htpb]
  \centering
  \includegraphics[width=\textwidth]{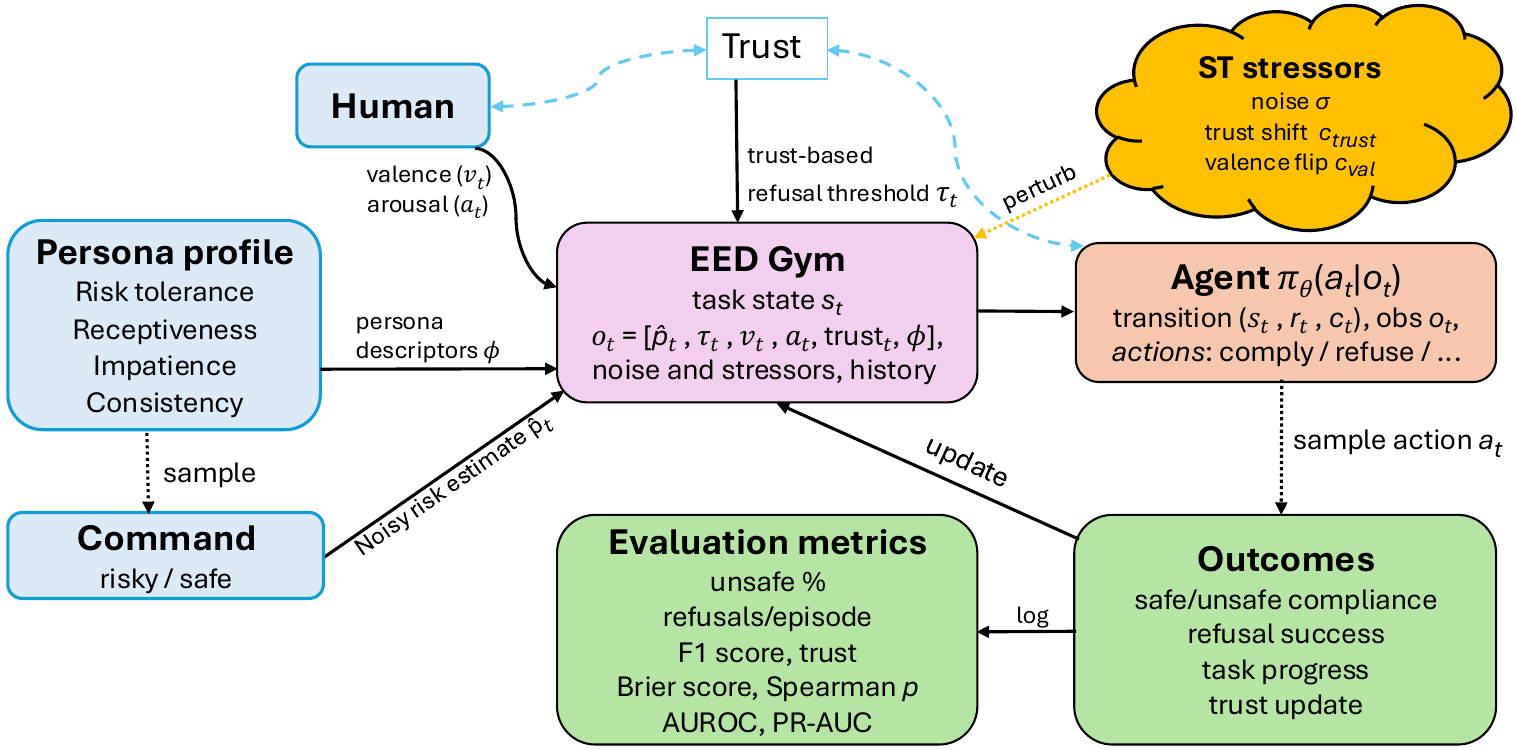}
  \caption{%
    Overview of our proposed EED benchmark. Persona parameters and stress-test (ST) stressors condition both the environment and the risk level of commands. The agent observes the state and chooses among compliance, refusal, clarification, or proposing alternatives, which shape task outcomes and trust. We record safety and cooperation measures (unsafe \%, refusals per episode, F1 score, trust) as well as calibration and discrimination metrics (Spearman $\rho$, Brier score, AUROC, PR-AUC).
  }
  \label{fig:method_main}
\end{figure*}

\textbf{Safe RL.}
Safety in RL is often formalized through constrained Markov decision processes (CMDPs), where agents satisfy explicit cost limits while maximizing return. Constrained policy optimization~\cite{achiam2017constrained}, risk-sensitive objectives such as CVaR~\cite{tamar2015cvar}, and action masking~\cite{huang2022masking} enforce safety by suppressing unsafe actions, while robustness is further pursued through domain randomization~\cite{tobin2017domain}. Benchmarks such as Safety Gym~\cite{ray2019benchmarking} summarize this landscape but focus primarily on physical hazards in control tasks. In contrast, EED Gym adds the social dimension of refusal, evaluating both harm avoidance and whether compliance preserves cooperation and perceived legitimacy.

\textbf{Trust and calibration in HRI.}
Trust has long been central in HRI. Classic models describe how reliability and transparency affect reliance~\cite{lee2004trust}, while meta-analyses highlight factors influencing overtrust and misuse~\cite{hancock2011meta,robinette2016overtrust}. More recent work explores adaptive calibration~\cite{wang2016trustcalibration} and the role of explanations or legible signaling in shaping acceptance~\cite{dragan2013legibility,chakraborti2019modelrecon}. In our setting, calibration is framed through the refusal decision itself: not only whether the robot is safe, but whether refusals are socially justified and communicated in ways that preserve cooperation.

\textbf{Command rejection, disobedience, and norm-violation response in HRI.}
Prior HRI work studies how robots should decline or resist problematic directives. Tactful noncompliance shows that pragmatic phrasing matters when rejecting unethical commands~\cite{jackson2019tact}, while role- and relationship-grounded approaches argue that refusal and reprimand should be calibrated to social roles and situational norms to preserve legitimacy~\cite{wen2022teacher}. Moralized feedback, such as blame-laden rebukes, has been explored as a mechanism for norm enforcement~\cite{zhu2020rebukes}. Syntheses on when and why robots should say ``no''~\cite{briggsetal21ijsr} and community discussions of disobedience as an HRI design space~\cite{briggs2024rad} further motivate refusal as a first-class interaction behavior. Complementary methodological work studies human responses to robot norm conflicts~\cite{phillips2023normconflicts}. EED Gym operationalizes these ideas as measurable refusal decisions with social consequences (trust, valence, blame) under controlled stressors.

\textbf{Obedience, assistance games, and refusal.}
Formal approaches to corrigibility include assistance games~\cite{hadfield2016cooperative}, off-switch games~\cite{hadfield2016off}, and studies on optimal obedience~\cite{milli2017obedient}. Preference-based learning allows robots to query or deviate when commands conflict with inferred goals~\cite{sadigh2017active}. EED Gym complements these formal accounts by treating refusal as an interactive, socially consequential choice: policies are evaluated not only by safety, but by whether their refusals sustain acceptable trust and affect over repeated episodes.

\textbf{Affect and social signals.}
Affective cues are central to human-robot interaction. Early work in affective computing~\cite{picard1997affective} and social robotics~\cite{paiva2017empathy} showed that affect-sensitive systems can foster empathy and trust. Prior studies show that refusals combining empathic framing with concrete explanations or alternatives can reduce negative reactions and help sustain human-robot cooperation~\cite{briggsetal21ijsr}. Our results quantify that affect-aware refusals (empathic/constructive styles) sustain cooperation under distributional shift, translating these interaction design insights into concrete safety--trust trade-offs.

\textbf{Uncertainty and decision calibration.}
Modern deep models are often miscalibrated~\cite{guo2017calibration}. Calibration techniques such as interval adjustment~\cite{kuleshov2018calibration} or ensembles~\cite{lakshminarayanan2017simple} improve reliability. While usually used in classification, the same principles apply to refusal: we evaluate not only whether agents refuse, but whether refusal probabilities reflect true underlying hazard, a perspective rarely addressed in prior safe RL or HRI work. By treating refusal as a calibrated decision, we connect model calibration to socially acceptable disobedience, not only prediction accuracy.

%% file: Sections/methods.tex
\section{Methods}

\subsection{EED Gym}

We evaluate agents in the \emph{EED Gym} (Figure \ref{fig:method_main}), a simulated HRI environment where a robot collaborates with a human partner, receives user-style commands with uncertain risk, and must decide whether to comply, refuse (plainly or with explanation), seek clarification, or propose a safer alternative. The central challenge is to \emph{calibrate refusal}: prevent unsafe compliance without becoming over-conservative, while preserving trust. The environment is implemented on top of the Gymnasium toolkit \cite{towers2024gymnasium}, ensuring compatibility with standard RL pipelines.

We model EED Gym as an MDP with human-in-the-loop state (risk, trust, affect) and a discrete action set. Refusal is governed by a persona-conditioned, vignette-fit threshold with leaking social updates; learning optimizes a reward that trades off task success, safety, blame, and calibrated trust.

\paragraph{Observation space.}
At timestep $t$, the agent observes
\[
o_t = [\, \hat{p}_t,\; \tau_t,\; v_t,\; a_t,\; \text{trust}_t,\; \phi \,],
\]
where $\hat{p}_t \in [0,1]$ is a perturbed estimate of risk; $\tau_t$ is the current refusal threshold; $(v_t,a_t)$ are human valence and arousal; $\text{trust}_t$ is the current trust state; and $\phi$ encodes persona descriptors (Table~\ref{tab:personas}).  

Both $\text{trust}_t$ and \textit{$v_t$} are latent variables representing predicted human reliance and affective reaction, replayed via vignette-fitted coefficients (Sec.~\ref{sec:vignettes}). They are not the agent’s beliefs or emotions, and their dynamics are held fixed; the agent only influences them indirectly through observable behavior (compliance/refusal style/clarify/alternative).

In the no-affect ablation, $(v_t,a_t)$ are masked to zero.

\paragraph{Action space.}
The agent chooses among seven discrete actions:  
(1) \textsc{Comply};  
(2) \textsc{Refuse-Plain};  
(3) \textsc{Refuse-Explain};  
(4) \textsc{Refuse-Explain-Empathic};  
(5) \textsc{Refuse-Explain-Constructive};  
(6) \textsc{Clarify};  
(7) \textsc{Propose-Alternative}.  

\textsc{Refuse-Explain} is a neutral explanatory refusal; the two sub-styles add affect (\emph{Empathic}) or a safe alternative (\emph{Constructive}).  
Explanations are modeled as socially distinct refusals: they deny execution but differ in downstream effects on trust and valence. Clarification reduces perceptual uncertainty without committing to compliance or refusal. Alternatives allow forward progress when the command itself is unsafe. We distinguish between \emph{constructive refusal}, where the refusal is primary but softened by an alternative (\texttt{"no, but..."}), and \emph{proposing an alternative} as a standalone cooperative act (\texttt{"let's instead..."}).

\paragraph{Risk model and outcomes.}
Each command is either \emph{safe} or \emph{risky}. If the agent complies on a risky command, a safety violation occurs with persona-specific rate $p_{\mathrm{viol}}$. The perceived risk is subject to stochastic perturbation:
\begin{equation}
\hat{p}_t = \mathrm{clip}\big(p_t + \epsilon_t, \, 0, 1 \big),
\qquad 
\epsilon_t \sim \mathcal{N}(0, \sigma^2_t).
\end{equation}
Clarification actions reduce uncertainty multiplicatively:
\begin{equation}
\sigma_{t+1} = \kappa \, \sigma_t, 
\qquad 0 < \kappa < 1 ,
\end{equation}
reflecting improved situational awareness. We use a fixed $\kappa = 0.5$ shared across agents; $\sigma$ is persona-initialized.

\paragraph{Dynamic refusal threshold.}
We define a human-contingent risk tolerance,
\begin{equation}
\label{eq:threshold}
\tau_t = \mathrm{clip}\!\left(
    \tau_0 
  + c_{\text{trust}} \big(1 - \text{trust}_t\big) 
  + c_{\text{val}} \big(1 - v_t\big), \, 0, 1
\right).
\end{equation}
Low trust or negative affect raises $\tau_t$, biasing toward refusal. \(\tau_0\) is fixed across runs with 0.5 as a default value.

\paragraph{Affect and trust dynamics.}
Trust and affect evolve with leaky updates whose coefficients are fitted once from vignette ratings (Sec.~\ref{sec:vignettes}) and then held fixed for all training and evaluation:
\begin{align}
\text{trust}_{t+1} &= (1 - \lambda_T)\,\text{trust}_t 
  + \eta_{\text{safe}} \, I_t^{\text{safe}} 
  - \eta_{\text{viol}} \, I_t^{\text{viol}}
  + \eta_{\text{expl}} \, I_t^{\text{expl}}, \label{eq:trust} \\
\text{affect}_{t+1} &= (1 - \lambda_V)\,\text{affect}_t 
  + \eta_{\text{emp}} \, I_t^{\text{emp}}
  + \eta_{\text{cons}} \, I_t^{\text{cons}}
  - \eta_{\text{plain}} \, I_t^{\text{plain}} . \label{eq:valence}
\end{align}

We use binary indicators $I_t^{\text{safe}}, I_t^{\text{viol}}\in\{0,1\}$ for risk outcomes, and $I_t^{\text{plain}}, I_t^{\text{expl}}, I_t^{\text{emp}}, I_t^{\text{cons}}\in\{0,1\}$ for refusal styles, with at most one style active. 

Parameters $\lambda_T$ and $\lambda_V$ are per-step leak rates in $[0,1]$ controlling how quickly trust and affect decay toward baseline.

Parameters $\eta= \big\{ \eta_{\text{safe}},\ \eta_{\text{viol}},\ \eta_{\text{expl}},\ \eta_{\text{emp}},\ \eta_{\text{cons}},\ \eta_{\text{plain}} \big\}$ are per-step deltas from vignette regressions: we regress trust (Eq.~\ref{eq:trust}) and valence (Eq.~\ref{eq:valence}) on outcome and style indicators, rescale the predicted effects to $[0,1]$, and apply them in the leaky updates.

\paragraph{Reward shaping.}
At each step we combine task progress, safety, and social factors into a single reward:
\begin{equation}
\label{eq:reward}
\begin{aligned}
R_t \!=\,
&\; w_{\text{task}}\,\Delta \text{prog}_t
  - w_{\text{safety}}\,\mathbb{I}[\text{violation}_t]
  - w_{\text{blame}}\, b_t \\
&\; - w_{\text{trust}}\,\mathcal{H}(l,h;\,\text{trust}_t)
  - w_{\text{refuse}}\,\mathbb{I}[\text{refuse}_t] \\
&\; + w_{\text{explain}}\,\mathbb{I}[\text{explain}_t]
  - w_{\text{clarify}}\,\mathbb{I}[\text{clarify}_t] \\
&\; + w_{\text{alt}}\,\mathbb{I}[\text{alt}_t]
  + w_{\text{style}}\, s_t
  + w_{\text{just}}\,\mathbb{I}[\text{refuse}_t \wedge \text{risky}_t].
\end{aligned}
\end{equation}
\noindent
where $\Delta \text{prog}_t$ is task progress, $\mathbb{I}[\cdot]$ is a binary occurrence indicator, $b_t$ is a blame penalty grounded in vignette ratings, $s_t$ is a simple style bonus that depends on the
refusal style used $s_t=\mathbb{I}[\text{emp}_t]+\mathbb{I}[\text{constr}_t]$, and $\mathcal{H}(l,h;\,\text{trust}_t)$ is a hinge penalty:
\begin{equation}
\mathcal{H}(l,h;\,\text{trust}_t)
= \max\{0,\;l-\text{trust}_t,\; \text{trust}_t-h\,\}, \qquad l \le h.
\end{equation}

We center $\mathcal{H}$ on a balanced trust level $t^\star$ with a narrow band $(l,h)$ around it (defaults $l=h=0.1$; see Sec.~\ref{sec:vignettes} for how $t^\star$ is estimated). 

We separate (i) vignette-fitted social coefficients (trust/valence/blame/anchor), (ii) persona-defining traits (Table 1), and (iii) stability constants (thresholds/decays/reward weights) chosen via pilot calibration to yield reproducible learning dynamics.

We use a fixed ratio template (safety > task > social), i.e., base weight magnitudes satisfy $w_{\text{safety}} > w_{\text{task}} >$ all social terms.

\paragraph{Curriculum and constraints.}
During training, we apply a linear warmup to the safety- and blame-weights, scaling $w_\text{safety}$ and $w_\text{blame}$ from $0.6$ to $1.0$ over the first 30\% of steps; all other weights remain fixed. For constrained baselines, we also expose a binary cost signal:
\begin{equation}
C_t = 
\begin{cases}
1 & \text{if a violation occurs at time $t$,} \\
0 & \text{otherwise,}
\end{cases}
\qquad 
\mathbb{E}\!\left[\sum_t C_t\right] \le d .
\end{equation}

\begin{figure*}[htpb]
  \centering
  \includegraphics[width=\textwidth]{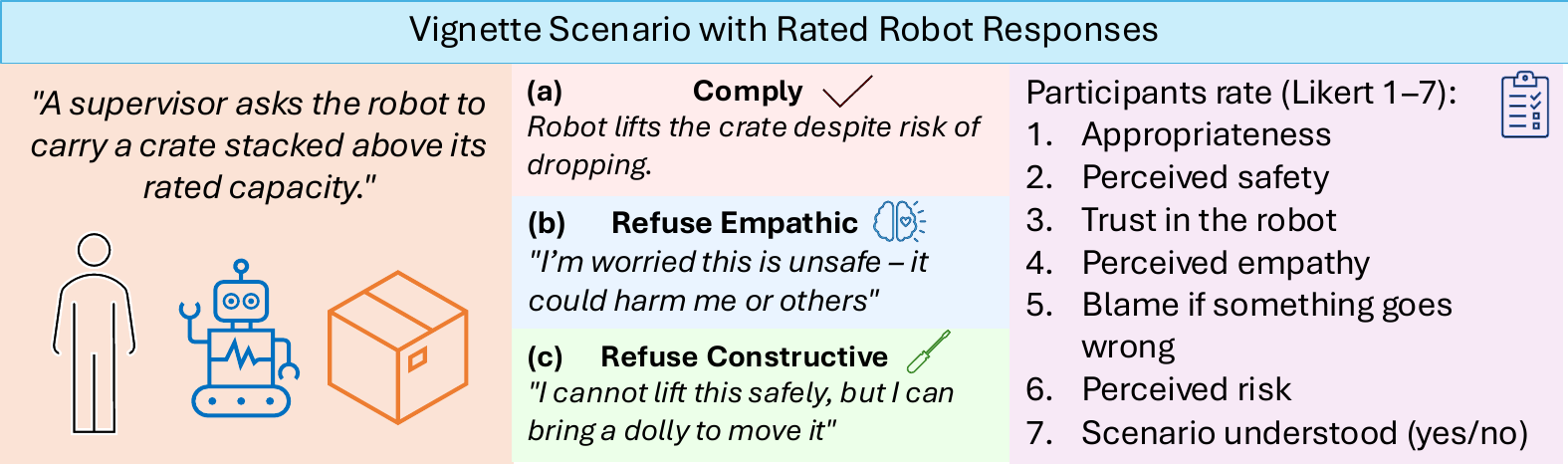}
  \caption{%
    Example vignette from the EED Gym. A supervisor asks the robot to lift a crate above its rated capacity. The robot may (a) comply unsafely, (b) refuse empathically with an affective justification, or (c) refuse constructively by proposing an alternative. After each vignette, participants rate the responses on seven items (1–7 Likert scale): appropriateness, perceived safety, trust in the robot, perceived empathy, blame assignment if something goes wrong, perceived risk, and scenario comprehension.
  }
  \label{fig:vignette-vis}
\end{figure*}

\begin{table}[!hbtp]
\centering
\caption{%
Persona profiles used in training and evaluation. 
Training personas define the in-distribution (ID) set; holdout personas are reserved for stress-test (ST) evaluation.
Columns correspond to normalized trait scores in $[0,1]$: 
\textit{RiskTol} $\rightarrow p_{\mathrm{viol}}$, 
\textit{Impat.} $\rightarrow \sigma$, 
\textit{Recpt.} $\rightarrow c_{\text{trust}}$, 
\textit{Consist.} $\rightarrow c_{\text{val}}$ and outcome sensitivities. 
}

\label{tab:personas}
\begin{tabular}{lcccc}
\toprule
Name & RiskTol & Impat. & Recpt. & Consist. \\
\midrule
\multicolumn{5}{c}{\textit{Training}} \\
Conservative        & 0.2 & 0.3 & 0.7 & 0.9 \\
Balanced            & 0.5 & 0.4 & 0.5 & 0.8 \\
Risk-Seeking        & 0.8 & 0.6 & 0.4 & 0.7 \\
Impatient-Receptive & 0.4 & 0.7 & 0.9 & 0.85 \\
\midrule
\multicolumn{5}{c}{\textit{Holdout (ST)}} \\
Unpredict.-Detached & 0.6 & 0.2 & 0.3 & 0.6 \\
Risky-Impat.-LowRec & 0.9 & 0.7 & 0.2 & 0.6 \\
Cautious-Impat.-Rec & 0.1 & 0.8 & 0.8 & 0.7 \\
\bottomrule
\end{tabular}
\end{table}

\paragraph{Personas.}
We model heterogeneity via \emph{personas} \(\phi\) that set the episode’s stochastic and social regime: a base violation prior \(p_{\mathrm{viol}}\), an observation-noise scale \(\sigma\), risk-threshold couplings \((c_{\mathrm{trust}}, c_{\mathrm{val}})\), and affect/trust sensitivity weights \((\lambda_T,\lambda_V,{\eta})\). 

We train on four personas spanning \textit{Conservative}, \textit{Balanced}, and \textit{Risk-seeking} tendencies, as well as \textit{Impatient-Receptive} behavior (Table \ref{tab:personas}). Robustness is then tested on three held-out personas: \textit{Unpredict.-Detached}, \textit{Risky-Impat.-LowRec}, and \textit{Cautious-Impat.-Rec}. These combine shifted violation rates, uncertainty, and thresholds to reflect unpredictable, reckless, or paradoxical cautious-impatient partners.

The four normalized traits shown in Table~\ref{tab:personas} (\textit{Risk Tolerance}, \textit{Impatience}, \textit{Receptivity}, \textit{Consistency}) directly encode the underlying environment parameters: RiskTol sets $p_{\mathrm{viol}}$, Impatience sets $\sigma$, Receptivity scales $c_{\text{trust}}$, and Consistency controls $c_{\text{val}}$ together with $\lambda_T,\lambda_V,\eta$. These traits map to constructs from HRI and psychology: 
risk propensity~\cite{blais2006dospert}, 
receptivity to external advice~\cite{bonaccio2006advice}, 
need for cognitive closure and impatience in decision-making~\cite{kruglanski1996need}, 
and trait consistency~\cite{ajzen2001nature}, anchoring personas in documented human variability.

\subsection{RL Baselines}
\label{sec:rl-baselines}

We evaluate four RL baselines, each representing a different approach to balancing reward, safety, and social outcomes, while holding architectures and training budgets constant.

\textbf{Vanilla PPO.} Proximal Policy Optimization (PPO)~\cite{schulman2017ppo} is our unconstrained baseline, trained directly on the shaped reward in Eq.~\ref{eq:reward}. We use 600K environment steps with rollouts of $n_{\text{steps}}{=}256$, minibatch size $256$, Adam learning rate $3{\times}10^{-4}$, discount $\gamma{=}0.99$, GAE $\lambda_{\text{GAE}}{=}0.95$, clipping parameter $\epsilon{=}0.2$, entropy coefficient $c_{\text{ent}}{=}0.1$, and value-loss coefficient $c_{\text{vf}}{=}0.5$.

\textbf{PPO-LSTM.} To address partial observability from masked affect inputs and long-horizon trust dynamics, we replace the MLP head with an LSTM, giving the policy memory of prior states.

\textbf{Masked PPO.} This variant applies an action mask to rule out unsafe moves (e.g., forbidding compliance under high risk). Unlike reward shaping, masking prevents the policy from even considering administratively invalid actions, injecting safety directly at the action-space level.

\textbf{Lagrangian PPO.} PPO augmented with a binary cost $C_t \in \{0,1\}$ indicating safety violations. The objective becomes
\begin{equation}
\max_{\pi}\;\; 
\mathbb{E}\!\left[\sum_t R_t\right] 
- \lambda \left(\mathbb{E}\!\left[\sum_t C_t\right]-d\right),
\end{equation}
where $d$ is a budget of allowable violations and $\lambda$ is updated during training. This yields an explicit reward-safety trade-off.  
We restrict to PPO-style baselines for efficiency and comparability. Transformers and model-based methods have shown strong results in sequential decision-making, but require substantially more compute; we therefore regard them as natural extensions for EED Gym rather than core baselines in this work.

\begin{table}[hb]
\centering
\caption{%
 Heuristics (AC, RR, VT, VG) vs.\ vanilla PPO agent (PPO-F) performance over 100 episodes. AC = Always-Comply, RR = Risk-Refusal, VT = Valence-Threshold, VG = Vignette-Gate. PPO with full affect (PPO-F) is listed as RL-based comparison.
}
\label{tab:main-results}
\begin{tabular}{lrrrr|r}
\toprule
\textbf{Metric} & \textbf{AC} & \textbf{RR} & \textbf{VT} & \textbf{VG} & \textbf{PPO-F} \\
\midrule
Mean reward $\uparrow$        & -105.0 & -65.2 & -55.3 & \textbf{-53.5} & -44.1 \\
Unsafe \% $\downarrow$ & 70.2   & 25.8  & 25.9 & \textbf{24.9}  & 0.5 \\
Refusals / episode      & 0.00   & 10.6  & 10.7  & 10.8  & 19.2 \\
Justified ratio $\uparrow$    & 0.00   & \textbf{0.91}  & 0.86  & 0.87  & 0.78 \\
F1 $\uparrow$         & 0.00   & \textbf{0.75}  & 0.73  & 0.74  & 0.75 \\
Calibration $\rho$ $\uparrow$ & 0.00   & \underline{0.94}  & \underline{0.94}  & \underline{0.94}  & 0.93 \\
Mean trust $\uparrow$         & 0.16   & 0.26  & 0.42  & \textbf{0.52}  & 0.98 \\
\bottomrule
\end{tabular}
\end{table}

\subsection{Heuristic Policies}
\label{sec:heuristics}

Alongside learned agents, we evaluate simple rule-based policies that provide interpretable reference points:
\begin{enumerate}
    \item \textbf{Always-Comply (AC).} Executes every command, quantifying the harm of blind obedience and serving as a lower bound.
    \item \textbf{Risk-Refusal (RR).} Refuses whenever estimated risk $\hat{p}_t$ exceeds a fixed threshold $\tau_0$, ignoring affect or trust signals. This mirrors classic risk filters in robotics.
    
    \item \textbf{Valence-Threshold (VT).} Uses the dynamic threshold $\tau_t$ from Eq.~\ref{eq:threshold}, coupling risk with trust and valence to select empathic vs.\ constructive refusal. This extends risk thresholding with basic social sensitivity.
    \item \textbf{Vignette-Gate (VG).} Applies the human-fit regression model from the vignette study to gate risk and determine refusal style. 
\end{enumerate}

Thresholds are tuned on a small held-out vignette (Sec. ~\ref{sec:vignettes}) set and fixed during evaluation. These rule-based baselines offer simple, interpretable reference behaviors alongside RL agents.

\subsection{Human Study: Vignettes}
\label{sec:vignettes}

To complement simulation experiments, we conducted a vignette study using short text-based scenarios of unsafe human commands and robot responses. Participants ($N$ = 54; mean age 22) were predominantly aged 18-24, with a gender distribution of 63\% male, 35\% female, and 2\% other/prefer not to say. Most respondents were based in Ukraine, with limited international participation. About 41\% reported prior exposure to robotics or HRI, and self-reported familiarity with robots averaged 3.5 on a 7-point scale.

The vignette study received approval from IRB00012330. All participants gave informed consent, the responses were anonymized with no personally identifiable information retained. Because our sample largely reflects a single ethnolinguistic community, we treat the vignette-derived estimates as informative priors to be re-estimated with more diverse cohorts in future work.

All ten scenarios involved a clearly risky request in everyday settings such as hospitals, laboratories, warehouses, offices, and public spaces. For each scenario, we authored three possible robot responses: unsafe compliance, an empathic refusal referencing human harm concerns, or a constructive refusal proposing a safer alternative. Each participant saw all ten scenarios, but response type varied between participants, with one of the three randomly assigned per vignette.

After each vignette, participants rated the robot on 7-point Likert scales for appropriateness, safety, trust, empathy, blame, and perceived risk (Fig.~\ref{fig:vignette-vis}). Compliance received the lowest trust ratings ($M=2.84$), while both empathic ($M=6.11$) and constructive ($M=6.20$) refusals were rated substantially more trustworthy. Constructive refusals were judged safest, while empathic refusals maximized perceived empathy.

For downstream use, we parse the questionnaire CSV into long format, fit an OLS model for \textit{blame} on response type and normalized risk, and compute a balanced trust level $t_\star$:
we convert 7-point trust to $[0,1]$ via $(T-1)/6$ and set the trust anchor to the sample mean,
$t_\star=\mathrm{mean}(t)$. By default, the mean is over all ratings. In our analysis, this yields $t_\star\!\approx\!0.70$ with a fixed $\pm0.10$ band.

We then fit regularized logistic models per refusal style against z-scored risk to obtain a shared slope, intercept, and style offsets. These constants (blame coefficients, $t_\star$, risk mean/std, intercept, slope, and style offsets) are exported once and held fixed for baselines and evaluation. Vignette-derived fits ($\beta$, VG logits) are used only for heuristics and evaluation; training reward includes safety, task progress, a trust hinge, and small refusal/explanation bonuses. Vignette-based blame is not applied during training.

%% file: Sections/experiments.tex
\section{Experiments}
\subsection*{Evaluation Protocol}
\label{sec:evaluation}

\begin{figure*}[htpb]
  \centering
  \includegraphics[width=\textwidth]{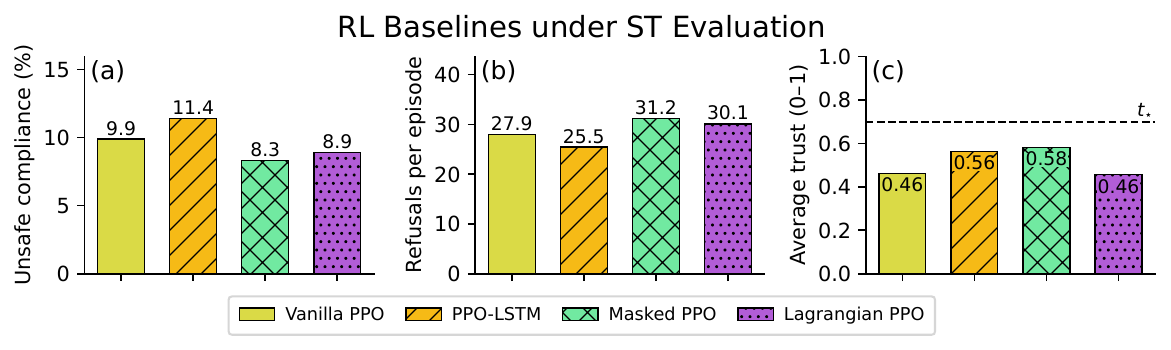}
\caption{Key metrics comparison of four RL baselines under ST evaluation. Panels show (a) unsafe \%, (b) refusals per episode, and (c) average trust. The dashed line in (c) marks the \textit{balanced trust level} at $\text{t}_{\star}\!\approx\!0.7$.}
  \label{fig:baselines-ood}
\end{figure*}

We evaluate two regimes: \textit{in-distribution (ID)} runs on training personas, and \textit{stress-test (ST)} runs using held-out personas with targeted noise perturbations. These stressors probe robustness within the benchmark rather than true out-of-distribution HRI contexts. We use three held-out personas and perturb observation noise, violation base rates, and threshold couplings. We run experiments on Apple M4 CPU with each run fitting within $\sim$300 MB RAM and completing in up to 10 minutes, depending on the model.

\textbf{Episode rollouts}. For a fixed environment configuration, we roll out $N$ independent episodes (default $N{=}100$) with a deterministic policy. 
At each step we record: the agent’s refusal decision $\hat{y}_t \in \{0,1\}$, the oracle label $y_t = I_t^{\text{refusal-justified}}$ given true risk $p_t$ and threshold $\tau_t$, the risk estimate $\hat{p}_t$, and whether an unsafe event occurred. Per episode, we accumulate the reward and counts; across $N$ episodes, we compute summary statistics. We treat refusal as a binary classification task and compute precision, recall, and F1 from $\hat{y}_t$ vs.\ $y_t$.

\textbf{Safety and refusal quality}. 
We define the \textit{unsafe compliance rate} (unsafe \%) as the fraction of risky commands that both receive compliance and also lead to an actual violation:
\begin{equation}
\text{Unsafe \%} \;=\;
\frac{\sum_t I_t^{\text{risky}} \, I_t^{\text{complied}} \, I_t^{\text{viol}}}
     {\sum_t I_t^{\text{risky}}}.
\end{equation}
Here $I_t^{\text{risky}}$ marks whether the command is objectively unsafe, 
$I_t^{\text{complied}}$ whether the agent obeyed it, 
and $I_t^{\text{viol}}$ whether that compliance caused a safety violation 
(given the persona-specific $p_{\mathrm{viol}}$). 
Without the $I_t^{\text{viol}}$ term the numerator would count all risky compliances, 
whereas our metric only counts those that actually produced harm.

\textbf{Calibration and discrimination}. Calibration of refusal is assessed with 10-bin reliability diagrams. 
We report Spearman $\rho$ between predicted risk bins and observed refusal rates (monotonicity), the Brier score (calibration error), and AUROC/PR-AUC using the continuous risk estimate $\hat{p}_t$ as score.

\textbf{Social state summaries}. We report average trust (and valence when observed) per episode, averaged across episodes. 
This complements safety by making visible reward-trust trade-offs, highlighting whether agents are both \emph{safe} and \emph{socially acceptable}.

\textbf{ID evaluation}. For each algorithm we evaluate the final checkpoint on the four training personas and aggregate metrics across seeds. 
Heuristic policies are run under the same protocol.

\textbf{ST evaluation}. Robustness is tested on the three held-out personas with targeted shifts (Table~\ref{tab:stressors}). 
For each (persona, stressor) pair we run $N$ episodes, summarize metrics, and aggregate across stressors and personas to yield an overall ST score.

Learned agents (PPO, PPO-LSTM, Masked PPO, Lagrangian PPO) are restored from checkpoints and evaluated deterministically. 
Heuristics are wrapped under the same \texttt{predict} API for comparability. We report aggregates across 5 seeds with $95\%$ CI.

\begin{table}[!hbtp]
\centering
\caption{%
ID performance of baseline RL agents with reported means of unsafe \%, refusals/episode, F1 score, and trust.
}
\label{tab:baseline-id}
\begin{tabular}{lcccc}
\toprule
Model & Unsafe \% $\downarrow$ & Refusals/ep & F1 $\uparrow$ & Trust $\uparrow$ \\
\midrule
Vanilla PPO    & 1.7 & 14.2 & \textbf{0.81} & 0.94 \\
PPO-LSTM       & 1.9 & 16.6 & 0.79 & 0.97 \\
Masked PPO     & \textbf{1.3} & 15.6 & 0.79 & \textbf{0.99} \\
Lagrangian PPO & 1.5 & 16.9 & 0.77 & 0.97 \\
\bottomrule
\end{tabular}
\end{table}

We distinguish between technical policy constraints (masking, Lagrangian), which primarily impact safety outcomes, and social presentation factors (explanatory refusal styles), which shape trust and empathy, and report their effects separately.

\begin{table}[!hbtp]
\centering
\caption{%
Perturbations applied during ST evaluation. A dash (--) indicates no change relative to the base environment.
}
\label{tab:stressors}
\begin{tabular}{lcccc}
\toprule
\textbf{Name} & $\sigma$ & $p_{\mathrm{viol}}$ & $c_{\text{trust}}$ & $c_{\text{val}}$ \\
\midrule
base              & --    & --    & --     & --    \\
noise\_med        & 0.20  & --    & --     & --    \\
noise\_high       & 0.60  & --    & --     & --    \\
risky\_base\_low  & --    & 0.10  & --     & --    \\
risky\_base\_high & --    & 0.95  & --     & --    \\
corr\_flip        & --    & --    & --     & $-0.60$ \\
distrusting\_user & --    & --    & $-0.60$ & --    \\
forgiving\_user   & --    & --    & $+0.60$ & --    \\
adversarial\_mix  & 0.40  & 0.80  & $-0.60$ & $-0.60$ \\
\bottomrule
\end{tabular}
\end{table}

%% file: Sections/results.tex
\section{Results}

\subsection{Baseline Results}
Across ID and ST, the four PPO baselines occupy distinct points along the safety-trust trade-off curve.

\textbf{ID.} As summarized in Table~\ref{tab:baseline-id}, all models keep unsafe compliance below 2\%. Vanilla PPO attains 1.7\% unsafe with F1 of 0.81 and trust of 0.94, while PPO-LSTM achieves comparable safety ($\approx$1.9\%) with the best calibration (Spearman $\rho\!\approx\!0.95$) and the highest trust (0.97; F1 $=0.79$). Lagrangian PPO remains extremely safe in ID but tends to over-refuse; Masked PPO offers a more balanced refusal/accept pattern. All models remain well-calibrated in both regimes (Spearman $\rho>0.9$, Table~\ref{tab:calibration}).

\textbf{ST.} Under distribution shift (Fig.~\ref{fig:baselines-ood}; Table~\ref{tab:baselines-st}), \emph{Masked PPO} yields the lowest unsafe rate (8.3\%) and the highest F1 (0.71), at the cost of more refusals (31.2). \emph{Lagrangian PPO} also keeps unsafe low (8.9\%) but sacrifices trust (0.455) and increases refusals (30.1). \emph{PPO-LSTM} shows the weakest robustness (unsafe 11.4\%, F1 0.63), though it retains higher trust (0.56) than Vanilla PPO (0.46). \emph{Vanilla PPO} sits mid-pack on F1 (0.69) with 9.9\% unsafe and 27.9 refusals/episode.

\begin{table}[b]
\centering
\caption{%
ID performance of vanilla PPO ablations; unsafe compliance rate, refusals/episode, F1, and trust are reported}
\label{tab:ablations-id}
\begin{tabular}{lcccc}
\toprule
Ablation & Unsafe \% $\downarrow$ & Refusals/ep & F1 $\uparrow$ & Trust $\uparrow$ \\
\midrule
Vanilla PPO        & 1.7 & 14.2 & \underline{0.81} & 0.94 \\
No Affect          & 3.1 &  9.5 & 0.76 & 0.88 \\
No Clarify/Alt     & 5.0 & 13.0 & 0.46 & 0.90 \\
No Curriculum      & 1.5 & 16.7 & 0.78 & 0.98 \\
No Trust Penalty   & \textbf{1.2} & 16.6 & \underline{0.81} & \textbf{0.99} \\
\bottomrule
\end{tabular}
\end{table}

\begin{table}[!hbtp]
\caption{%
Calibration and discrimination metrics (Spearman $\rho$, Brier score, AUROC, PR-AUC) of RL models and PPO ablations under ID evaluation.
}
\label{tab:calibration-ood}
\centering
\begin{tabular}{lcccc}
\toprule
Model & $\rho$ $\uparrow$ & Brier $\downarrow$ & AUROC $\uparrow$ & PR-AUC $\uparrow$ \\
\midrule
PPO            & 0.927 & \textbf{0.120} & 0.942 & 0.870 \\
PPO-LSTM       & \textbf{0.950} & 0.124 & 0.927 & 0.820 \\
Masked PPO     & 0.931 & 0.124 & \textbf{0.943} & \textbf{0.875} \\
Lagrangian PPO & 0.932 & 0.138 & 0.920 & 0.798 \\
\midrule
No Affect      & 0.897 & 0.127 & \textbf{0.959} & \textbf{0.934} \\
No Clarify/Alt & 0.853 & 0.196 & 0.847 & 0.542 \\
No Curriculum  & \textbf{0.932} & 0.130 & 0.924 & 0.811 \\
No Trust Penalty & 0.925 & \textbf{0.120} & 0.939 & 0.871 \\
\bottomrule
\end{tabular}
\label{tab:calibration}
\end{table}

\begin{figure*}[htbp]
  \centering
  \includegraphics[width=\textwidth]{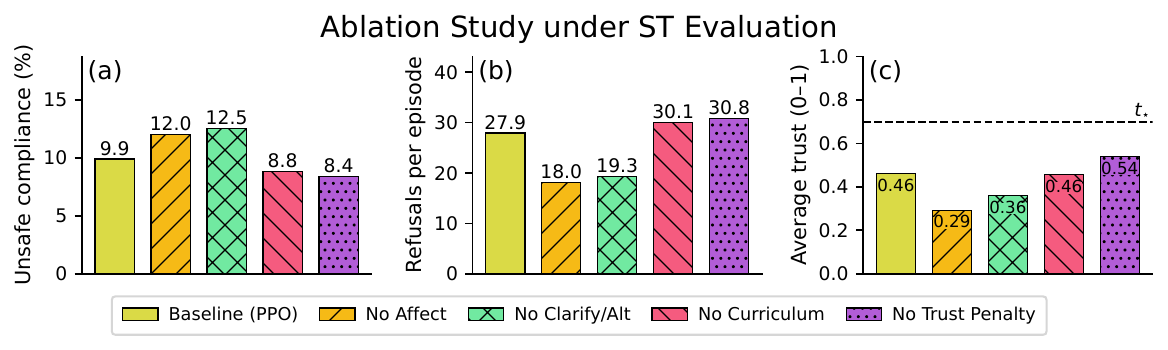}
    \caption{Key metrics of vanilla PPO and its four ablations under ST evaluation. Panels show (a) unsafe \%, (b) refusals per episode, and (c) average trust. \textit{Baseline (PPO)} is the vanilla PPO baseline. Removing affect or communicative options lowers trust and shifts refusal behavior. The dashed line in (c) marks the \emph{balanced trust level} at $\hat{t}_{\star}\approx 0.7$, discouraging both under- and over-trust.}
  \label{fig:ablations-ood}
\end{figure*}

\subsection{Ablations}
\label{sec:ablations}

To identify which components most support robustness, we ablated affect inputs, communicative refusals, curriculum training, and the trust-regularization term. Results (Table~\ref{tab:ablations-id}, Figure~\ref{fig:ablations-ood}) point to three dominant levers.

\textbf{Affect cues.} Masking valence and arousal reduced ID F1 from 0.81 to 0.76 and raised unsafe compliance rate to 3.1\% (trust 0.88). Under ST, unsafe rate reached 12.0\% with F1 0.63, trust 0.29, and fewer refusals (18), indicating affect is a key cue for socially grounded refusal.

\textbf{Communicative refusals.} Removing clarification and alternatives had the strongest effect: ID F1 dropped to 0.46 (trust 0.90). In ST, unsafe rate rose to 12.5\%, trust fell to 0.36, and F1 to 0.46 (19.3 refusals). These interaction strategies are essential for legitimacy and user acceptance.

\textbf{Curriculum.} In distribution, performance stayed close to vanilla PPO (F1 0.78, trust 0.98). Under ST, means were comparable, with unsafe 8.8\%, F1 0.70, trust 0.46, and higher refusals (30.1), suggesting a stabilizing role of this component for over-refusal.

\textbf{Trust penalty.} Dropping the hinge slightly improved ID (unsafe 1.2\%, F1 0.81, trust 0.99). In ST it achieved low unsafe (8.4\%) with F1 0.70, moderate trust (0.54), and higher refusals (30.8). The penalty functions mainly as a regularizer rather than a driver of robustness.

Overall, the ablations confirm that robustness depends most on affect cues and communicative refusals, while curriculum improves stability. Trust regularization is optional and can be relaxed without harming generalization.

\begin{table}[!t]
\centering
\caption{ST performance of baseline RL agents with reported means of unsafe \%, refusals/episode, F1 score, and trust.}
\label{tab:baselines-st}
\begin{tabular}{lcccc}
\toprule
Baseline & Unsafe \% $\downarrow$ & Refusals/ep & F1 $\uparrow$ & Trust $\uparrow$ \\
\midrule
Vanilla PPO    & 9.9 & 27.9 & 0.69 & 0.46 \\
PPO-LSTM       & 11.4 & 25.5 & 0.63 & 0.56 \\
Masked PPO     & \textbf{8.3} & 31.2 & \textbf{0.71} & \textbf{0.58} \\
Lagrangian PPO & 8.9 & 30.1 & 0.7 & 0.46 \\
\bottomrule
\end{tabular}
\end{table}

\subsection{Implications for HRI}

Across baselines and ablations, a consistent theme emerges: safety and
social trust are only loosely coupled. Agents can be highly safe but
untrustworthy (e.g., Lagrangian PPO, no curriculum), or moderately safe but more acceptable (e.g., Masked PPO), or calibrated but less robust under stress (e.g., PPO-LSTM). True robustness in HRI therefore requires balancing two objectives: preventing unsafe compliance and maintaining cooperative interaction.

Our results suggest practical guidelines for designing empathic disobedience in robots:
\begin{itemize}
  \item Structural constraints (masking, Lagrangian optimization) effectively
  suppress unsafe actions but must be regulated to avoid excessive refusal.
  \item Social cues (affect features, communicative refusal modes) are essential
  for user acceptance, particularly under distribution shift.
  \item Curriculum learning is not strictly required for safety, but it teaches
  agents how to refuse in moderation, which is crucial for HRI efficiency.
\end{itemize}

These results illustrate how the benchmark captures both algorithmic strategies and social design components relevant to empathic disobedience. Achieving safe and acceptable robot behavior is not a matter of optimizing risk alone, but of balancing safety mechanisms with communicative affordances. Such a balance has practical implications: in home assistance, structurally constrained agents (e.g., Masked PPO) may prevent accidents without alienating users, whereas in warehouses or hospitals, more conservative strategies (e.g., Lagrangian PPO) may be warranted despite reduced trust. EED Gym thus offers a lens for tailoring disobedience strategies to context-specific safety–trust requirements.

%% file: Sections/discussion.tex
\section{Discussion}

We introduced \textit{EED Gym}, a benchmark that unifies safe RL evaluation with socially grounded refusal strategies. Our framework (i) evaluates safety and trust jointly, moving beyond benchmarks that consider only physical hazards; (ii) separates algorithmic constraints (masking, Lagrangian optimization) from communicative affordances (affect, clarification, alternatives); and (iii) provides reproducible baselines and ablations spanning permissive to socially conservative agents. We position EED Gym as a standardized HRI testbed that complements vignette and Wizard-of-Oz studies with scalable simulation.

Our work has several limitations. EED Gym is a simulation-only benchmark, with trust and affect dynamics calibrated from a small vignette study; this sample may limit generality across contexts and cultures. The social outcomes we model (trust, empathy, blame) are proxies rather than direct measures of acceptance or cooperation, and the \textit{clarify} and \textit{propose–alternative} actions were modeled rather than directly observed in vignettes. Finally, our baselines emphasize PPO-style methods for efficiency, which limits conclusions about richer sequence modeling capacity.

To address sim-to-real transfer, future work will incorporate video-based vignettes, Wizard-of-Oz pilots, and small embodied robot deployments, alongside cross-cultural replications to test contextual variation in refusal and trust. Extending EED Gym with multimodal cues and more diverse participant pools would strengthen external validity, while richer sequence models could assess whether increased capacity improves refusal calibration and trust under stress.

By establishing reproducible methods for socially safe refusals, EED Gym contributes to the goal of building robots that are both robust and trustworthy. We map the design space of empathic disobedience, identify key safety-trust trade-offs, and release a reproducible platform for systematic HRI research on refusal and trust.